\pdfoutput=1
\documentclass[11pt]{article}

\usepackage[]{acl}

\usepackage{times}
\usepackage{latexsym}

\usepackage[T1]{fontenc}
\usepackage[utf8]{inputenc}
\usepackage{enumitem}
\usepackage{microtype}
\newcommand{\multiwoz}{MultiWOZ}

\newcommand{\gpt}{GPT-2}

\newcommand{\bert}{BERT}
\newcommand{\bertlarge}{BERT-Large}
\newcommand{\roberta}{RoBERTa}
\newcommand{\robertalarge}{RoBERTa-Large}
\newcommand{\reinforce}{REINFORCE}
\newcommand{\soloistparg}{\textsc{Soloist+PARG}}
\newcommand{\soloistoa}{\textsc{Soloist-OA}}
\newcommand{\soloistth}{\textsc{Soloist-TH}}

\newcommand{\soloist}{\textsc{Soloist}}
\newcommand{\soloiststeach}{\textsc{Soloist$_{\texttt{S}}$+Teach}}
\newcommand{\slsoloist}{\textsc{SL-Soloist}}
\newcommand{\slsoloistteach}{\textsc{SL-Soloist+Teach}}
\newcommand{\soloistthteach}{\textsc{Soloist-TH+Teach}}

\newcommand{\slagent}{\textsc{SL-Agent}}
\newcommand{\modelp}[1]{\soloist{$_{\texttt{#1}}$}}

\newcommand{\ie}[0]{\emph{i.e., }}

\newcommand{\eg}[0]{\emph{e.g., }}

\newcommand{\etc}[0]{\emph{etc.}}

\newcommand{\RN}[1]{%
	\textup{\lowercase\expandafter{\it \romannumeral#1}}%
}
\usepackage{graphicx}
\usepackage{caption}
\usepackage{amsmath}
\usepackage{threeparttable}
\usepackage{booktabs}
\usepackage{tabularx}
\usepackage{adjustbox}
\usepackage{algorithm}
\usepackage{algorithmic}
\usepackage{multirow}
\usepackage{wasysym}
\title{Toward Self-learning End-to-End Task-Oriented Dialog Systems}
\author{Xiaoying Zhang$^{1}$, Baolin Peng$^{2}$, Jianfeng Gao$^{2}$, Helen Meng$^{1}$ \\
    $^{1}$The Chinese University of Hong Kong, Hong Kong\\
    $^{2}$Microsoft Research, Redmond\\
    \{zhangxy, hmmeng\}@se.cuhk.edu.hk  \\ 
    \{bapeng, jfgao\}@microsoft.com}
\begin{document}
\maketitle
\begin{abstract}

End-to-end task bots are typically learned over a static and usually limited-size corpus. However, when deployed in dynamic, changing, and open environments to interact with users, task bots tend to fail when confronted with data that deviate from the training corpus, \ie out-of-distribution samples. In this paper, we study the problem of automatically adapting task bots to changing environments by learning from human-bot interactions with minimum or zero human annotations. 
We propose \slagent{}\footnote{\textsc{\b{S}elf-\b{L}earning Agent}.}, a novel self-learning framework for building end-to-end task bots. \slagent{} consists of a dialog model and a pre-trained reward model to predict the quality of an agent response. It enables task bots to automatically adapt to changing environments by learning from the unlabeled human-bot dialog logs accumulated after deployment via reinforcement learning with the incorporated reward model. Experimental results on four well-studied dialog tasks show the effectiveness of \slagent{} to automatically adapt to changing environments, using both automatic and human evaluations. We will release code and data for further research.

\end{abstract}

\section{Introduction}
\begin{figure}[t]
\centering
\includegraphics[width=0.95\columnwidth]{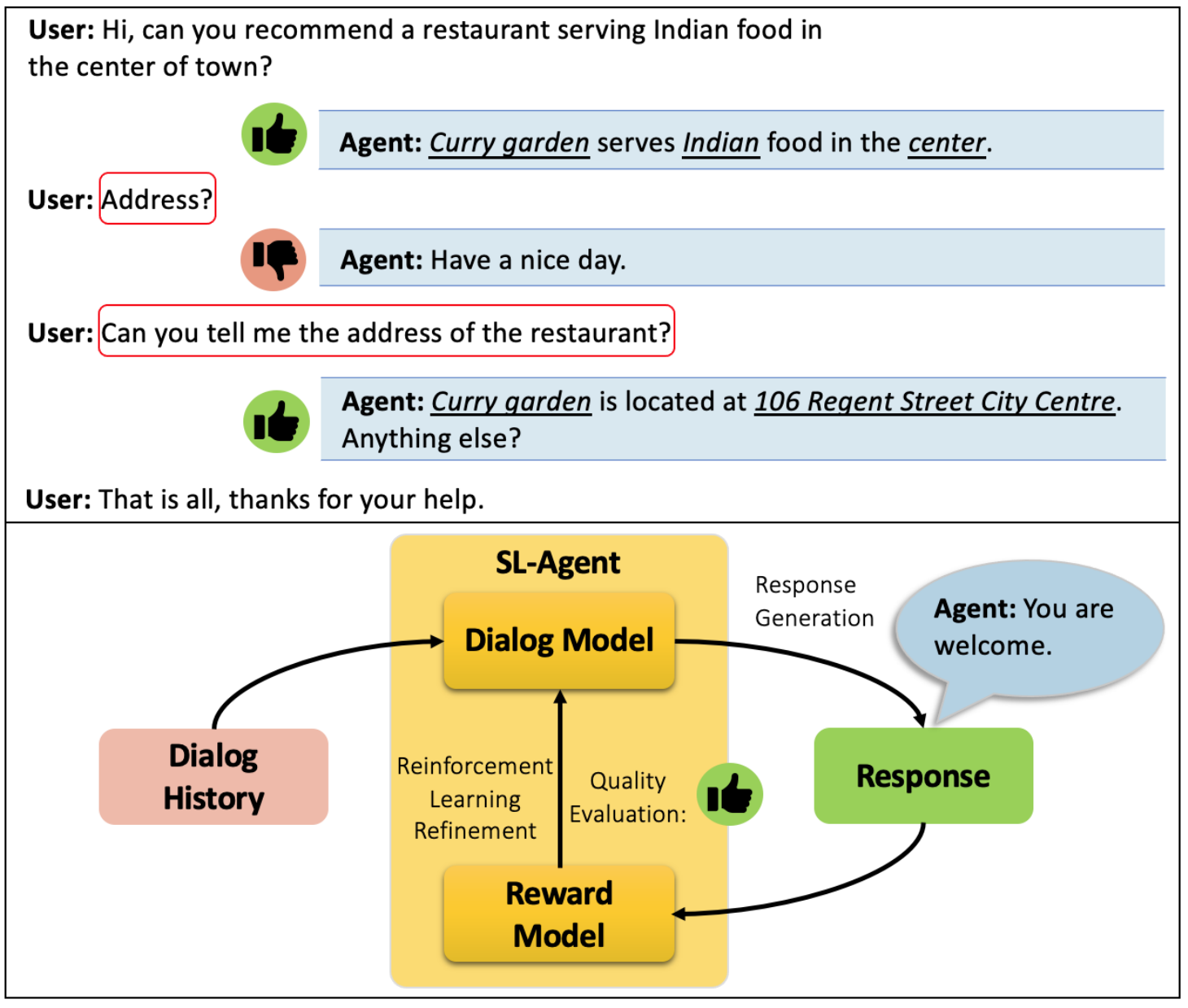}
% \caption{The proposed \slagent{} framework. The dialog history on top, is fed into the dialog model for belief state tracking and response generation. Then the reward model judges the quality of a response and yields a reward score to guide the refinement of the dialog model using reinforcement learning. }
\caption{Illustration of the proposed \slagent{} with a human-bot dialog example. 
$(\RN{1})$ The human-bot dialog example, containing an inappropriate response related to unseen user behaviors (upper part). $(\RN{2})$ Demonstration of the refining process in \slagent{} with the exhibited dialog example (lower part).
%The bot fails to respond with address due to language variations at first, but gives the correct response in the following attempt. These human-bot interactions contain useful information from which the task bot can learn to improve its performance using proposed \slagent{}.
}

\label{fig:running_example}
\end{figure}
% data-driven approach - bad
The most common approach of building end-to-end task-oriented dialog systems is to train neural models to imitate human behaviors in fixed task-specific annotated corpora \cite{gao2018neural,zhang2020task}. Existing state-of-the-art approaches usually adopt Pre-trained Language Models (PLMs) \cite{peng2020soloist, Ham2020e2e, hosseini2020simple} to build end-to-end dialog systems. However, these data-driven approaches assume an independent and identically distributed (IID) data setting\footnote{Assume the same user behaviors at deployment as in the training stage.}, \ie a static environment\footnote{Environment is the Agent’s world in which it lives and interacts.}, and usually exhibit a tendency of failure, when confronted with out-of-distribution (OOD) examples in real-world scenarios, \ie changing environments.

In the context of task-oriented dialog systems, changing environments are quite common and arise from the following two aspects:
$(\RN{1})$ \textit{unseen user behaviors} -- real users may query with unseen language patterns and unknown user goals (\ie unseen slot values and dialog flows) of the designated tasks outside the pre-built training corpora \cite{liu2018dialogue,peng2020raddle}. For example, real users may query entities in the database but not covered by the training examples.
%For example, real users may query with more diverse expressions than crowd-sourced templates or query entities in the database but not covered by the training examples.
%real user queries may contain unseen language patterns and slot values of the designated tasks than pre-built training corpora. 
%(\ie various language expressions for querying the same item, in which utterance to query the item \cite{shah2018bootstrapping})
% As shown in the upper part of Figure \ref{fig:running_example}, after deployment, when user queries casually about address, the system fails to provide address in the second response, but gives it in the third response, when user queries in a detailed way (similar to the training examples). % (\ie language variation)
%Furthermore, repeatedly querying about the address demonstrates the policy variation, as in the train set, there is no failed attempts.
%For example, the address querying in the Table-booking task of restaurants is usually expressed in a detailed way (``Can you tell me the address of the restaurant?") in the training set. However, in real-world scenarios, some users tend to ask casually (``Address?"), and the system tends to fail in such cases.
$(\RN{2})$ \textit{task definition extensions} -- dialog systems need to handle new functions or new tasks as user and business requirements evolve, \ie add new slot types \cite{lipton2018bbq, Gasic2014IncrementalOA}. For example, a restaurant bot designed for the table-booking service may also encounter queries about delivery service after deployment. These human-bot interactions accumulated after deployment are cheap, dynamic and contain useful information \cite{hancock2019learning}, \ie unseen user behaviors are related to the training examples and the probabilistic dialog model may generate appropriate responses. As shown in the upper part of Figure \ref{fig:running_example}, when user queries casually about address, the system fails to provide address in the second response, but gives it in the third response, when user queries in a detailed way (similar to the training examples).
Therefore, rather than merely imitating human behaviors in a fixed corpus, task bots are desired to spontaneously learn from the interactions with real users, progressively improve and adapt after being deployed in dynamic and constantly changing environments.
% for task bots to learn to improve its performance
% build a task bot that can progressively improve and adapt after being deployed in the dynamic and constantly changing environments.

% To achieve this goal, a promising and efficient attempt is to directly learn from the interactions with real users \cite{shukla2020conversation}, which are cheap to collect and largely reflect the real-world scenarios, containing useful information for task bots to learn to improve its performance.
% To achieve this goal, a promising and efficient attempt is to directly learn from the human-bot interactions accumulated after deployment \cite{shukla2020conversation}, which are cheap, dynamic, and contain useful information for task bots to learn to improve its performance.
% \cite{shukla2020conversation}, which are cheap to collect, and contain potentially useful information for task bots to learn to improve its performance.
%the human-bot interactions accumulated after deployment are usually abundant, task-specific, dynamic, but contain potentially useful information to improve the bot

% There are several attempts to leverage human-bot interactions to improve task bots in changing environments. For example, \citet{liu2018dialogue, shah2018bootstrapping, dai2020learning} propose to query humans for adequate feedback or corrections to improve task bots. However, largely depending on human annotations or user feedback can be costly, and sometimes users are unwilling to give any feedback. In addition, how to leverage human-bot interactions to adapt task bots to new tasks 
% is rarely explored.
There are several attempts to leverage human-bot interactions to improve task bots in changing environments. For example, \citet{liu2018dialogue, shah2018bootstrapping, dai2020learning} propose to query humans for adequate feedback scores or annotations. However, it relies on human annotations or user feedback, which can be costly and sometimes users are unwilling to give any feedback. In addition, these works center on dialog policy optimization or retrieval-based task bots. Automatically adapting task bots to changing environments is imperative for end-to-end dialog model yet under-explored. Furthermore, these works usually omit task definition extensions. 

In this paper, we propose \slagent{}, a novel self-learning framework for building end-to-end task bots in a more realistic changing environment setting with minimum or zero human annotations. It consists of a neural dialog model and a pre-trained reward model, where the dialog model generates responses and the reward model judges the quality of agent responses.
Specifically, we devise a data augmentation strategy to construct positive and negative examples based on the given dialog training corpus to endow the reward model with the capability to judge the quality of responses for unlabeled human-bot dialog logs.
The bot (including dialog model and reward model) is first trained with the same available training data, then deployed to converse with real users and collect human-bot dialog logs. 
After that, as shown in the lower part of Figure \ref{fig:running_example}, the bot is refined with the unlabeled human-bot dialog logs via reinforcement learning, where the response quality is judged by the reward model. In this way, the bot can automatically adapt to unseen user behaviors, without extra human annotations. 
Regarding the problem of extensions in task definitions, machine teaching is utilized to correct representative failed dialogs with minimum human annotations to provide necessary instructions on how to handle new functions. After that, the bot quickly adapts to new functions through the self-learning procedure. 

Our contributions are summarized as below:

\begin{itemize}\setlength{\itemsep}{0pt}
% \item We propose a new research problem: ``how to enable a task bot to adapt to changing environments with minimum to zero human annotations''.

\item We propose a new research problem \ie how to enable task bots to automatically adapt themselves to changing environments by learning from interactions with minimum or zero human annotations.

\item We propose a novel self-learning framework \slagent{} that equips with a pre-trained reward model trained by the devised data-augmentation strategy to build generative end-to-end task bots in a realistic changing environment setting, with minimum or zero human annotations. 

% \item We devise a data augmentation approach to build a pre-trained reward model with the capability to judge the quality of responses for the unlabeled human-bot dialog logs.

\item We conduct comprehensive experiments on four datasets to demonstrate the effectiveness of \slagent{} for enabling automatic adaptation to changing environments by learning from the unlabeled human-bot dialog logs using both automatic and human evaluations.
% \item We demonstrate that \slagent{} enables automatic adaptation to changing environments on four dialog tasks, by learning from unlabeled human-bot dialog logs via reinforcement learning with the pre-trained reward model.
% and the results demonstrate that \slagent{} framework enables task bots to automatically adapt to changing environments by learning from the unlabeled human-bot dialog logs accumulated after deployment.
% We demonstrate that \slagent{} enables automatic adaptation to changing environments on four dialog tasks, by learning from unlabeled human-bot dialog logs via reinforcement learning with the pre-trained reward model, using both automatic and human evaluations.
% \item In a challenging domain extension setting, we show that, enhanced by machine teaching, \slagent{} effectively handles new functions with minimum human corrections. 

\end{itemize}

\section{Related Work}

% \paragraph{Active Learning}

\paragraph{RL for Dialog Policy Learning}
Reinforcement learning has been widely applied to dialog systems for policy optimization. \citet{young2013pomdp, peng2018deep,peng2017composite,liu2017iterative,Gasic2014IncrementalOA,tseng2021transferable} formulate dialog policy learning as a sequential problem and use REINFORCE \cite{williams1992simple} and/or Q-learning \cite{watkins1992q} to optimize the dialog policy. \slagent{} utilizes a similar REINFORCE algorithm but focuses on generative end-to-end optimization.

% \citet{wen2017latent, zhao2019rethinking} utilize RL to directly optimize reward scores defined by BLEU score and success rate for dialog systems. \citet{tseng2021transferable} proposes to jointly model dialog agent and user simulator. However, these works focus on dialog policy learning in a static environment, and we propose to learn from human-bot interactions to automatically adapt to changing environments via reinforcement learning.

% \slagent{} utilizes a similar REINFORCE algorithm but focuses on generative end-to-end optimization.

% % Several works apply RL to optimize against ground truth responses using automatic evaluation metrics , or joint dialog agent and user simulator policy optimization based on known user goals \cite{tseng2021transferable}. 

% Different from the works mentioned above, \slagent{} focuses on refining with unlabeled human-bot dialog logs, which may contain previously unseen user behaviors and uncertain user goals, \ie a more realistic changing environment setting. 

%Therefore, the aforementioned approaches cannot be directly compared within this paper. 
\paragraph{Adapting to Changing Environments for Dialog Systems} 
Several attempts have been made to deal with changing environments after deployment. \citet{rajendran2019learning,dai2020learning} propose to learn from the human-bot interactions but requires lots of human corrections. \citet{shah2018bootstrapping,liu2018dialogue,gavsic2011line,Gasic2014IncrementalOA} propose to learn from human-bot interactions via reinforcement learning based on the queried human feedback scores after each dialog. To reduce the efforts of querying humans, \citet{su2016line} introduces a session-level Bi-LSTM reward model trained with extra pre-collected classification corpus to predict the task success of each dialog. 
Nevertheless, session-level reward model may underestimate the quality of responses in single dialog turns. Different from the works mentioned above, \slagent{} leverages a turn-level pre-trained reward model built on the given dialog corpus using the devised data augmentation approach and focuses on generative end-to-end dialog systems. Another line of research is using data-augmentation methods to generate diverse user behaviors during the training stage \cite{gao2020paraphrase,li2020coco}. Additionally, \citet{madotto2020continual,liu2021domain} continually collect extra labeled data to train task bots but aim to overcome the catastrophic forgetting problem, which is a different research topic (\ie continual learning) from our paper.

\section{\slagent{}}

% \begin{figure}[t]
% \centering
% \includegraphics[width=0.95\columnwidth]{Figures/training_pipeline.pdf}
% \caption{Training pipeline of the proposed \slagent{}.}
% \label{fig:pipeline}
% \end{figure}
\begin{figure}[t]
\centering
\includegraphics[width=0.95\columnwidth]{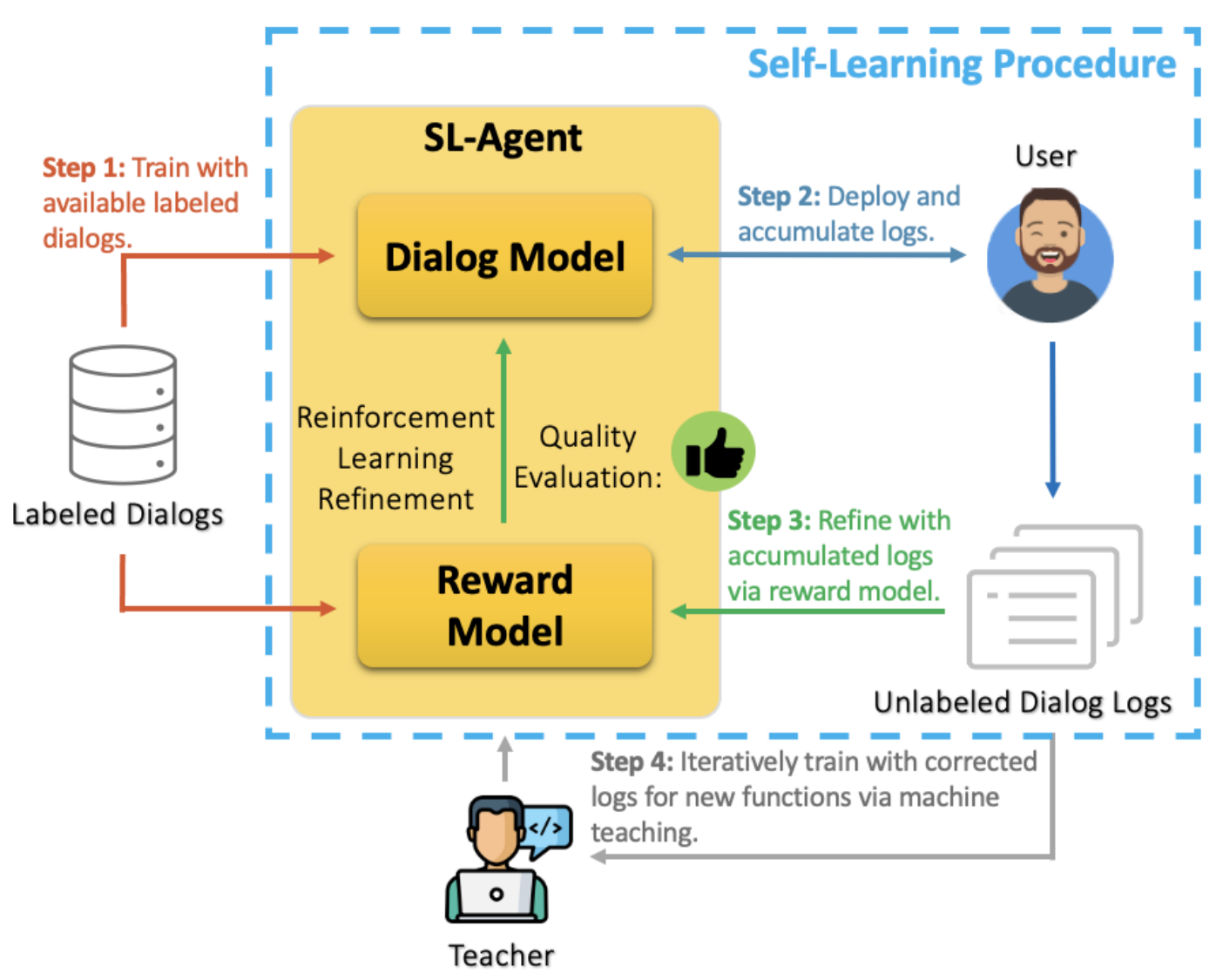}
\caption{Training pipeline of the proposed \slagent{}.}
\label{fig:pipeline}
\end{figure}
\subsection{Overview}
As depicted in Figure \ref{fig:pipeline}, \slagent{} contains two components: (i) a dialog model for generating responses (Section \ref{dialog_model}); (ii) a pre-trained reward model for judging the quality of agent responses and outputting reward scores to guide the refinement of the dialog model (Section \ref{reward_model}). Specifically, \slagent{} operates in the following steps: (i) First, the bot (both dialog model and pre-trained reward model) is fine-tuned with the same available annotated task-specific dialogs. (ii) Then, the bot is deployed online to converse with users and accumulate unlabeled human-bot dialog logs. (iii) Next, the dialog model is refined with these human-bot dialog logs via reinforcement learning, using the reward scores from the reward model (Section \ref{rl_refine}). $(\RN{4})$ For task definition extensions, machine teaching is utilized to correct representative failed dialogs to provide instructions on how to handle new functions (Section \ref{mt}). After that, the bot further improves through the self-learning procedure.
% where the response quality is judged by 
\subsection{Dialog Model}
\label{dialog_model}
\slagent{} is a general framework that is compatible with any generative end-to-end dialog models \cite{peng2020soloist,Ham2020e2e,hosseini2020simple}. In this paper, we employ \soloist{} \cite{peng2020soloist}, a pre-trained end-to-end dialog model, resulting in an agent termed \slsoloist{}\footnote{In this paper, \slagent{} refers to the proposed framework and \slsoloist{} is an instance of it, which utilizes \soloist{} as its dialog model.}.

We briefly review \soloist{} for completeness. \soloist{} formulates the end-to-end dialog generation as a sequence generation problem, by sequentially concatenating the inputs and outputs of 4 dialog modules (\ie NLU, DST, POL, NLG) in a typical dialog system. Each dialog turn is represented as:

\begin{equation}
    \boldsymbol{x}=(\boldsymbol{s}, \boldsymbol{b}, \boldsymbol{c}, \boldsymbol{r}),
    \label{eqa:x}
\end{equation}

\noindent where $\boldsymbol{s}$ is the entire dialog history, $\boldsymbol{b}$ is the annotated belief state, $\boldsymbol{c}$ refers to DB state fetched from database, and $\boldsymbol{r}$ is the delexicalized agent response. \soloist{} employs a Transformer-based model with parameters $\boldsymbol{\theta_D}$ to characterize the sequence generation probability $p_{\boldsymbol{\theta_{D}}}(\boldsymbol{x})$. Initialized with \gpt{} \cite{radford2019language}, the model is pre-trained on large-scale annotated dialog corpora, and then fine-tuned with limited task-specific dialogs. 

\paragraph{Synthetic Dialog Construction.}

To identify user behaviors with unseen slot values, we propose to synthesize dialog examples by exhausting database (DB) values and substitute corresponding slot values of in the training set. Specifically, for each dialog turn $\boldsymbol{x}$, we replace slot values in the utterances and user goal with corresponding new values of the randomly sampled DB entry.

\subsection{Reward Model}
\label{reward_model}
% motivation function

%  with new language patterns and uncertain user goals (\ie slot values)
The human-bot dialog logs accumulated after deployment may contain previously unseen user behaviors with unseen language patterns and unknown user goals. To enable the dialog model to identify these new types of user inputs to which the previously trained system cannot respond appropriately, we propose a reward model that judges the quality of an agent response through a reward score (a positive reward for an appropriate response, a negative reward for an inappropriate response). 

We formulate the quality evaluation problem as a binary classification task. Dialog responses are jointly determined by the dialog history, generated belief state, and fetched DB state. Therefore, given the training data $\mathcal{D}$ (annotated with belief states and DB states), we build a turn-level reward model $R$, which is parameterized by a Transformer $\boldsymbol{\theta_{R}}$ with the input dialog turn sequence $ \boldsymbol{x}$, defined as Equation \ref{eqa:x} to characterize the classification probability: $p_{\boldsymbol{\theta_{R}}}(\boldsymbol{x}) = p_{\boldsymbol{\theta_{R}}}(\boldsymbol{s}, \boldsymbol{b}, \boldsymbol{c}, \boldsymbol{r})$.

% Furthermore, each dialog turn of these dialog logs contains the generated belief state by the dialog model and fetched DB state, and intuitively the . Therefore, the reward model is expected to evaluate the quality of the generated belief state and response.
%% reward model input architecture
% Specifically, we propose a turn-level reward model $R$, which is a Transformer parameterized by $\boldsymbol{\theta_{R}}$ with inputs $ \boldsymbol{x}$ defined as Equation \ref{eqa:x}.

% \begin{equation}
%     y = p_{\boldsymbol{\theta_{R}}}(\boldsymbol{x}) = p_{\boldsymbol{\theta_{R}}}(\boldsymbol{s}, \boldsymbol{b}, \boldsymbol{c}, \boldsymbol{r}).
% \end{equation}

% To endow the reward model with the capability of predicting the response quality, we train the reward model based on Noise Contrastive Estimation \cite{gutmann2010noise}.

The reward model $R$ is trained using contrastive objective to discriminate between an appropriate response (\ie positive example $\boldsymbol{x}$) and an inappropriate response (\ie negative example $\boldsymbol{\hat{x}}$), given the dialog history. Specifically, for each dialog turn, we construct several positive examples $\left\{\boldsymbol{x}_{m}\right\}_{m=1}^{M}$ and negative examples $\left\{\boldsymbol{\hat{x}_n}\right\}_{n=1}^{N}$ based on the sequence $\boldsymbol{x}$, to add the relevance of real-world scenarios and endow the reward model with the capability of evaluating the response quality. Then we apply a binary classifier on top of the output sequence representation from the Transformer to discriminate between a positive example $\boldsymbol{x}$ ($y=1$) and a negative example $\boldsymbol{\hat{x}}$ ($y=0$). The training objective for a single example in the training set $\mathcal{D}$ is defined as: %in the training set $\mathcal{D}$ is
% \begin{equation}
%     \mathcal{L}_{\boldsymbol{\theta_{R}}} = \sum_{m=1}^{\mathcal{M}}\log \left(p_{\boldsymbol{\theta_{R}}}(\boldsymbol{x}_m)\right)+ \sum_{n=1}^{\mathcal{N}}\log \left(1-p_{\boldsymbol{\theta_{R}}}\left(\boldsymbol{\hat{x}_n}\right)\right),
% \end{equation}
\begin{equation}
\begin{aligned}
    \mathcal{L}_{\boldsymbol{\theta_{R}}} = &y \sum_{m=1}^{\mathcal{M}}\log \left(p_{\boldsymbol{\theta_{R}}}(\boldsymbol{x}_m)\right) \\
    +& (1-y)\sum_{n=1}^{\mathcal{N}}\log \left(1-p_{\boldsymbol{\theta_{R}}}\left(\boldsymbol{\hat{x}_n}\right)\right),
\end{aligned}
\end{equation}
% \noindent where $|\mathcal{M}|$ is the number of positive examples, $|\mathcal{N}|$ is the number of negative examples. 
% The full training objective over the entire training set $\mathcal{D}$ is $\mathcal{L}_{\boldsymbol{\theta_{R}}}(\mathcal{D}) = $

\paragraph{Positive Examples.}
For each dialog turn, we consider two kinds of user utterances: $(\RN{1})$ the original user utterance in the training set $\mathcal{D}$, to identify the appropriate response to the user behavior; $(\RN{2})$ the paraphrased user utterances generated based on the original user utterance using back translation \cite{edunov2018understanding}, to enhance the ability of reward model for identifying user behaviors with diverse language patterns. 
%(We use the Transformer-based machine translation checkpoint (English-German) \cite{edunov2018understanding} to generate 5 paraphrased user utterances for each dialog turn, based on empirical analysis of translation quality.)

\paragraph{Negative Examples.}
Based on the analysis on 200 human-bot dialog logs collected from the evaluation platform of DSTC8 Track 1 challenge \cite{li2020results}\footnote{These human-bot dialog logs contain the evaluation scores and comments from Amazon Mechanical Turks.}, we summarize 5 types of dialog turns that have inappropriate responses (in Appendix \ref{sec:neg}). Then, for each dialog turn in the training data $\mathcal{D}$, we construct negative examples $\boldsymbol{\hat{x}}$ (in brackets) according to these 5 types: 

\begin{itemize}\setlength{\itemsep}{0pt}
    \item \textbf{Repetition} The dialog model failed to understand the user's repeated query and generated the same response twice. (Repeating the response from the previous turn.)
    % \textit{repeat the  response from the previous turn});
    \item \textbf{Inconsistency} The dialog model generated an incoherent response. (Randomly sampling a response from the dataset $\mathcal{D}$ to replace the original response .)
    % (\textit{randomly replaced response, response generated by untrained dialog model});
    \item \textbf{Partial Information} The dialog model partially understood user request and answered incompletely. (For those user utterances with multiple slots request, randomly dropping a slot answer in the original response.)
    % (\textit{truncated response});
    \item \textbf{Non-fluency} The dialog model generated a non-fluent response. (Randomly repeating some word tokens in the original response.)
    \item \textbf{Misunderstanding} The dialog model generated the incoherent belief state and response. (Randomly sampling a belief state and response from the dataset $\mathcal{D}$ to replace the original belief state and response.)
    
    % % \item \textbf{Partial Understanding} The dialog model failed to understand the user's intent, and generated incoherent belief states. We generate a as a negative example. (\textit{randomly replaced belief}).
\end{itemize}

% \begin{itemize}\setlength{\itemsep}{0pt}
%     \item original user utterances, 
%     \item high-quality paraphrased user utterances generated by back translation \cite{edunov2018understanding}.
% \end{itemize}
% In order to For each dialog turn, we use back translation \cite{edunov2018understanding} to augment the user utterances with high-quality paraphrases, \ie diverse language patterns.

To boost the model performance with limited annotated task-specific corpora, we propose to follow the pre-training and fine-tuning paradigm to build the reward model, \ie pre-train the reward model using large-scale annotated heterogeneous dialog corpora, then fine-tune the pre-trained reward model with annotated task-specific data using the same training objective. The pre-training corpora is Schema dataset \cite{rastogi2019towards}.

%In addition, we use back translation \cite{edunov2018understanding} to augment the training examples with paraphrased user behaviors to further boost the performance.

\subsection{Refine with Reinforcement Learning}
\label{rl_refine}

% During the self-learning procedure, the dialog model is refined with these unlabeled, noisy dialogs, where the quality of agent responses is judged by the reward model. 
The interactions between the agent and users can be modeled as a sequential decision problem. As such, the dialog model can be refined via the \reinforce{} algorithm \cite{williams1992simple}. The policy is the trained dialog model $p_{\boldsymbol{\theta_D}}(\boldsymbol{x})$, the initial state is the dialog history $\boldsymbol{s}$, and the action space corresponds to the vocabulary set $\mathcal{V}$. The reward perceived by the dialog model is $R\left(\boldsymbol{s}, \boldsymbol{b}, \boldsymbol{c}, \boldsymbol{r} \right)$ from the reward model. The parameters $\boldsymbol{\theta_D}$ are updated by maximizing the cumulative reward score. The refining procedure is described in detail as follows:

For each RL episode, we randomly sample a dialog turn with dialog history and delexicalized response. We run the dialog model to generate belief state $\boldsymbol{\hat{b}}$, based on the input dialog history sequence $\boldsymbol{s}$. At each time step $t$, we sample a token $\hat{b}_{t}$ according to the model distribution, where the logits' distribution of the model is first filtered using Nucleus (top-p) filtering \cite{holtzman2019curious}, then redistributed via softmax function. Then we retrieve DB state $\boldsymbol{\hat{c}}$ from the database using $\boldsymbol{\hat{b}}$, and sample the delexicalized response sequence $\boldsymbol{r}$ following same sampling procedure, based on the token sequence $(\boldsymbol{s}, \boldsymbol{\hat{b}}, \boldsymbol{\hat{c}})$. Note that the delexicalized response is given as part of the input. Then we feed the concatenation of dialog history $\boldsymbol{s}$, generated belief state $\boldsymbol{\hat{b}}$, retrieved DB state $\boldsymbol{\hat{c}}$ and the response $\boldsymbol{r}$, i.e. $(\boldsymbol{s}, \boldsymbol{\hat{b}}, \boldsymbol{\hat{c}}, \boldsymbol{r})$ into the reward model $p_{\boldsymbol{\theta_{R}}}(\boldsymbol{x})$ to obtain the reward score $R(\boldsymbol{s},\boldsymbol{\hat{b}}, \boldsymbol{\hat{c}},\boldsymbol{r})$. The positive reward is 1, negative reward is -1. The training objective for a single example is represented as:
% $(R\boldsymbol{(s,b,c,r)}\rightarrow 1)$
\begin{equation}
\begin{aligned}
\mathcal{L}_{\theta_D}=-\sum_{t=1}^{T_{\hat{b}}} \log p_{\boldsymbol{\theta_D}}\left(\hat{b}_{t} \mid \hat{b}_{<t}, \boldsymbol{s}\right) \times R\boldsymbol{(s, \hat{b},\hat{c},r)} \\
    - \sum_{t=1}^{T_{r}} \log p_{\boldsymbol{\theta_D}}\left(r_{t} \mid r_{<t}, \boldsymbol{\hat{b}}, \boldsymbol{\hat{c}}, \boldsymbol{s}\right)
    \times R\boldsymbol{(s, \hat{b},\hat{c},r)}, \label{eqa:RL_ob}
\end{aligned}
\end{equation}
\noindent where the length of generated belief state and input delexicalized response are $\boldsymbol{T_{\hat{b}}}$, $\boldsymbol{T_{r}}$, respectively. Algorithm \ref{alg:Framwork} (in Appendix \ref{sec:alg}) summarizes the self-learning-based RL refining framework for refining the dialog model. 

\subsection{Minimum annotations via Machine Teaching} 
\label{mt}
To handle the queries about new functions in additional dialog turns, we need to introduce new slot-value pairs, action templates, \etc (An example is in Appendix \ref{sec:task_extension}.) Machine teaching is an efficient approach to training task bots \cite{simard2017machine, williams2017demonstration}. In this paper, we implement machine teaching via Conversational Learner (CL) \cite{shukla2020conversation}. The teaching process is conducted in three steps: $(\RN{1})$ The trained task bot is deployed online to fulfill the given goals by interacting with real users, leaving a handful of human-bot dialog logs. $(\RN{2})$ Human experts select a few representative failed dialogs to construct training examples with new functions by adding new action templates, introducing new slot-value pairs, correcting inappropriate responses and annotations (\ie belief states). $(\RN{3})$ The deployed task bot (\ie both dialog model and reward model) is trained on these training examples to handle new functions.

\section{Experiments}
% In this section, we first describe how we design evaluations on changing environments. Then we introduce the experiments we conduct on four well-studied dialog tasks using both automatic and human evaluation. 
\begin{table}[t]
  \centering
  \begin{threeparttable}
  \fontsize{7}{8}
  \selectfont
    \begin{tabular}{lcccc}
        \toprule
        {Domain}&{\texttt{Attraction}}&{\texttt{Train}}&{\texttt{Hotel}}&{\texttt{Restaurant}}\cr 
        \midrule
        \#Train&50&50&50&50\cr
        \#Valid&50&50&50&50\cr
        \#Test&100&200&200&200\cr
        \bottomrule
    \end{tabular}
  \end{threeparttable}
  \caption{Data statistics of four single-domain dialog datasets \cite{peng2020soloist,budzianowski2018multiwoz}.}
  \label{tab:data_statistics}
\end{table}

\begin{table*}[!t]
    \fontsize{7}{7}
    \selectfont
  \centering
  \begin{threeparttable}
    \begin{tabular}{lcccccccccccc}
    \toprule  
    \multirow{2}{*}{Model} &
    \multicolumn{3}{c}{\texttt{Attraction}}&\multicolumn{3}{c}{\texttt{Train}}&\multicolumn{3}{c}{\texttt{Hotel}}&\multicolumn{3}{c}{\texttt{Restaurant}}\cr  
      \cmidrule(lr){2-4} \cmidrule(lr){5-7} \cmidrule(lr){8-10} \cmidrule(lr){11-13}
      &$\mathtt{Inform} $&$\mathtt{Success}$&$\mathtt{BLEU}$&$\mathtt{Inform}$&$\mathtt{Success}$&$\mathtt{BLEU}$&$\mathtt{Inform}$&$\mathtt{Success}$&$\mathtt{BLEU}$&$\mathtt{Inform}$&$\mathtt{Success}$ &$\mathtt{BLEU}$ \cr 
    % \midrule
    % \multicolumn{14}{c}{Offline Preparation} \cr 
    \midrule
    \modelp{5} & 27.00&14.00&4.07&72.73&32.32&5.43&25.00&3.50&2.93&26.50&2.00&4.71\cr
    \modelp{S} & 60.00&33.00&8.14&73.74&54.55&6.94&56.00&29.50&7.05&62.50&41.50&7.33\cr
    \soloistparg{} & 60.00&32.00&8.83&75.25&56.06&8.45&58.00&29.00&7.71&64.00&42.00&9.17\cr
    % \multicolumn{14}{c}{Online Improvement} \cr
    \midrule
    \soloistoa{} & 61.00&36.00&8.66&74.75&55.05&7.58&56.50&29.00&7.14&64.50&42.50&8.56\cr
    \slsoloist{} &\textbf{64.00}&\textbf{40.00}&\textbf{8.99}&\textbf{75.76}&\textbf{61.62}&\textbf{10.97}&\textbf{60.50}&\textbf{39.50}&\textbf{8.34}&\textbf{75.00}&\textbf{44.50}&\textbf{10.60}\cr
    \soloistth{} &66.00&41.00&9.01&77.27&62.87&10.70&60.00&42.50&9.82&70.50&46.00&11.76\cr
    \midrule
    \modelp{50} &86.00&65.00&12.90&80.81&64.65&9.96&74.50&43.50&8.12&81.00&55.50&12.80\cr
    \bottomrule  
    \end{tabular}
  \end{threeparttable}
  \caption{End-to-end evaluation results on four tasks. The forth to sixth rows indicate the results of refining with 45 simulated (unlabeled) human-bot dialog logs, based on \modelp{S}. \modelp{50} is quoted from \citet{peng2020soloist}. (\slsoloist{} significantly outperforms all baselines in mean with p<0.01 based on Combined.)
%   End-to-end evaluation results on four tasks. \modelp{5} is trained with 5 examples. \modelp{S} refers to training with synthetic dialogs constructed from the 5 training examples. The third to fifth rows indicate the results of refining with 45 simulated human-bot dialogs using session-level reward from online activate reward model, \slagent{}, and turn-level human feedback score, respectively. \modelp{50} is quoted from \citet{peng2020soloist}. (\slsoloist{} significantly outperforms all baselines in mean with p<0.01 based on Combined.)
  %\pbl{several typos: 4 four cite a paper for \modelp{Rephrase} Paraphrase Augmented Task-Oriented Dialog Generation}
  }
  \label{tab:results_4task}
\end{table*}
\begin{table*}[!t]
    \fontsize{7}{7}
    \selectfont
  \centering
  \begin{threeparttable}
    \begin{tabular}{lcccccccccccc}
    \toprule  
    \multirow{2}{*}{Model} &
    \multicolumn{3}{c}{\texttt{Attraction}}&\multicolumn{3}{c}{\texttt{Train}}&\multicolumn{3}{c}{\texttt{Hotel}}&\multicolumn{3}{c}{\texttt{Restaurant}}\cr  
     \cmidrule(lr){2-4} \cmidrule(lr){5-7} \cmidrule(lr){8-10} \cmidrule(lr){11-13} &$\mathtt{Inform} $&$\mathtt{Success}$&$\mathtt{BLEU}$&$\mathtt{Inform}$&$\mathtt{Success}$&$\mathtt{BLEU}$&$\mathtt{Inform}$&$\mathtt{Success}$&$\mathtt{BLEU}$&$\mathtt{Inform}$&$\mathtt{Success}$ &$\mathtt{BLEU}$ \cr 
    \midrule
    \modelp{S} & 60.00&33.00&8.14&73.74&54.55&6.94&56.00&29.50&7.05&62.50&41.50&7.33\cr
    \soloistoa{} & 63.00&34.00&8.66&77.78&55.05&8.13&58.50&30.00&7.08&63.00&42.00&10.03\cr
    \slsoloist{} & \textbf{70.00}&\textbf{36.00}&\textbf{8.68}&\textbf{78.28}&\textbf{60.10}&\textbf{9.06}&\textbf{62.00}&\textbf{33.50}&\textbf{7.39}&\textbf{70.00}&\textbf{45.00}&\textbf{10.93}\cr
    \soloistth{} &68.00&40.00&9.01&76.77&62.63&9.55&62.50&35.50&7.83&70.50&47.50&11.36\cr
    \bottomrule  
    \end{tabular}
  \end{threeparttable}
  \caption{Automatic evaluation results on four tasks in Real-Scenario Setting. The first row refers to previously reported \modelp{S}. 
  The last three rows refer to refining with 30 real (unlabeled) human-bot dialog logs based on \modelp{S}. 
%   The last three rows refer to refining with 30 unlabeled, real human-bot dialogs using reward from \cite{su2016line}, \slagent{} and turn-level human feedback score, respectively. 
  (\slsoloist{} significantly outperforms all baselines in mean with p<0.01 based on Combined.)
  %\pbl{typos: 4 four; remove $_{Real}$}
  }
  \label{tab:policy_improve}
\end{table*}

\subsection{Experimental Setup}

We validate the efficiency and flexibility of proposed \slagent{} on four different end-to-end dialog tasks using \multiwoz{} single-domain dialog datasets \cite{budzianowski2018multiwoz}, reorganized by \citet{peng2020soloist}. Data statistics are shown in Table \ref{tab:data_statistics}.
Based on above datasets, we construct two settings to represent the changing environments -- \textbf{Setting I} for unseen user behaviors and \textbf{Setting II} for task definition extensions.

\paragraph{Implementation Details.}
To implement the proposed reward model, we conduct experiments with several Transformer-based models and \gpt{} \cite{radford2019language} (enhanced with auxiliary generation task) shows better performance than others. Therefore, we implement proposed reward model using \gpt{}-117M and the multi-task training objective. Full details are in Appendix \ref{sec:imple_details}.

\paragraph{Automatic Evaluation Metrics.} 
We report the results using the same automatic evaluation metrics following \citet{budzianowski2018multiwoz}: 
$(\RN{1})$ $\mathtt{Inform}(\%)$ evaluates whether the agent returns an appropriate entity. 
$(\RN{2})$ $\mathtt{Success}(\%)$ judges whether the agent correctly answers all requested attributes.
$(\RN{3})$ $\mathtt{BLEU}(\%)$ measures the word overlap of the generated response against human response. 
$(\RN{4})$ $\mathtt{Combined}(\%)$ assesses the overall quality, which is defined as: $\mathtt{Combined}$ = ($\mathtt{Inform}$ + $\mathtt{Success}$) $\times$ 0.5 + $\mathtt{BLEU}$.

\paragraph{Human Evaluation Metrics.}

Following the same evaluation protocol in the DSTC9 Track 1 challenge 
\cite{gunasekara2020overview}, we conduct human evaluations to judge the agent quality.
For each dialog session, Amazon Mechanic Turkers are presented with a goal and instructions, then they are required to converse with agent to achieve the goal via natural language. At the end of each dialog session, Turks are required to assess the overall dialog quality using the following five metrics:
$(\RN{1})$ $\mathtt{Success\ w/o\ g}(\%)$ judges whether the agent completes the task.
$(\RN{2})$ $\mathtt{Success\ w/\ g}(\%)$ judges whether the agent completes the task and provides matched slot values against the database record. 
$(\RN{3})$ $\mathtt{Understanding}$(1-5) measures the understanding correctness of user utterances. 
$(\RN{4})$ $\mathtt{Appropriateness}$(1-5) indicates the appropriateness, naturalness, and fluency of an agent response. 
$(\RN{5})$ $\mathtt{Turns}$ reports the average number of dialog turns for successful dialog sessions.

\paragraph{Compared Methods.}

To demonstrate the effectiveness of the proposed reward model in \slagent{}, we use \soloist{} as the dialog model to compare the performance of different methods. %for illustrating the effectiveness of \slagent{}.
\begin{itemize}\setlength{\itemsep}{0pt}
    \item \modelp{5} is trained with 5 labeled dialogs, randomly sampled from the train set.
    \item \modelp{S} is trained using synthetic dialogs constructed from the 5 labeled dialogs used for training \modelp{5}.
    \item \soloistparg{} is trained on \modelp{S} with paraphrased dialogs \cite{gao2020paraphrase,edunov2018understanding} constructed from the 5 labeled dialogs, \ie data-augmentation baseline for adapting to unseen user behaviors.
    \item \soloistoa{} is refined with unlabeled human-bot dialog logs based on \modelp{S} using the session-level reward of task success from online activate reward model (trained using the same 5 labeled dialogs as \modelp{5}) and partially queried session-level human feedback score \cite{su2016line}.
    \item \slsoloist{} (Ours) is refined with unlabeled human-bot dialog logs based on \modelp{S} using the pre-trained reward model in \slagent{}, which is fine-tuned using the same 5 labeled dialogs as \modelp{5}. Machine teaching is not utilized by now\footnote{To better demonstrate the self-learning capability of \slagent{}, machine teaching is only used in the setting of task definition extensions. However, machine teaching can be optionally used to update the bot for better performance in the setting of unseen user behaviors.}.
    \item \soloistth{} is refined with unlabeled human-bot dialog logs based on \modelp{S} using queried turn-level human feedback score, which is an upper bound.
    \item \modelp{50} is trained with whole 50 labeled dialogs, which can be regarded as the result of sufficient human corrections, \ie the highest bound. 
    (Details are shown in Appendix \ref{sec:exper_details}.)
\end{itemize}

\subsection{Results of Setting \uppercase\expandafter{\romannumeral1} - Unseen User Behaviors}

% 
% \footnote{Building a user simulator is inapplicable in our changing environment setting. $(\RN{1})$ It is difficult to build reliable user simulators. Building agenda-based user simulators requires sophisticated human expertise for designing rules. $(\RN{2})$ Building model-based user simulators requires sufficient labeled data. Furthermore, model-based user simulators merely imitate expert behaviors in the training corpus, cannot provide user behaviors that are unseen from task bots.}
\paragraph{Simulation Evaluation Setup.} Deploying a trained agent to interact with real human users and collect dialog logs is labor-intensive and costly for experimental purposes. Hence, we construct a setting to simulate unseen user behaviors. We randomly sample 5 dialogs from the training set as labeled data to train a task bot (\ie both dialog model and reward model). Note that the remaining 45 dialogs contain unseen user behaviors with unseen language patterns and unknown user goals. Hence, it is applicable to simulate unseen user behaviors by modifying the remaining 45 dialogs as unlabeled imperfect human-bot dialog logs (through adding noise, \ie corrupting responses\footnote{Note that the associated labels of belief states are not used. Construction details are in Appendix \ref{sec:simu_hb}.}). These 45 unlabeled human-bot dialog logs are further used for refining \modelp{S}, resulting in \soloistoa{}, \slsoloist{}, \soloistth{}. This simulation setting allows us to perform a detailed analysis of the reward model in \slagent{} without much cost and easily reproduce the experimental results.

\paragraph{Simulation Evaluation Results.} The end-to-end evaluation results on four different tasks are presented in Table \ref{tab:results_4task}. \modelp{S} significantly outperforms \modelp{5} over all evaluation metrics on all tasks, which shows the effectiveness of the proposed synthetic dialog construction for identifying user behaviors with unseen slot values. \slsoloist{} outperforms \soloistparg{} over all the metrics, which demonstrates the higher efficiency of directly learning from human-bot dialog logs. We observe that \slsoloist{} outperforms \soloistoa{} by a large margin, and achieves comparable performance with \soloistth{} (refining with turn-level human feedback score, \ie the upper bound). This shows the strong capability of the turn-level pre-trained reward model in \slagent{} for predicting the quality of responses. We conjecture that our proposed reward model trained with the proposed data-augmentation strategy is more robust to unseen user behaviors and thus ports richer useful information to dialog models. The results verify the vast potential of the proposed \slagent{}, allowing the bot to automatically adapt to unseen user behaviors without extra human annotations. Results of further policy improvement are shown in Appendix \ref{sec:policy_impr}.
%Compared to \modelp{50} (\ie the result of sufficient human corrections), \slsoloist{} (initialized with only 5 labeled dialogs) achieves impressive performance through automatically learning from 45 unlabeled, noisy human-bot dialog logs using \slagent{}.
\paragraph{Real-Scenario Evaluation Setup.} Simulation setting allows effortless experimental studies to validate the effectiveness of the reward model in \slagent{}. However, the results are likely biased. Therefore, in the real-scenario setting, we deploy \modelp{S} online and recruit human users to converse with it. We collect 30 real (unlabeled) human-bot dialog logs to refine \modelp{S}, resulting in the agent \soloistoa{}, \slsoloist{}, \soloistth{}.
\paragraph{Real-Scenario Evaluation Results.} The evaluation results on four tasks are shown in Table \ref{tab:policy_improve}. We observe that \slsoloist{} refined using the reward model in \slagent{} outperforms other methods over all evaluation metrics on all tasks. Furthermore, \slsoloist{} achieves comparable performance with \soloistth{}, even achieves better performance on certain metrics. We conclude that the results of real-scenario evaluation and simulation evaluation are consistent, confirming that \slsoloist{} enables effective self-learning after deployment by learning from interactions.

\subsection{Results of Setting \uppercase\expandafter{\romannumeral2} -- Task Definition Extensions}

\begin{table}[!t]
  \centering
  \begin{threeparttable}
  \fontsize{9}{10}
  \selectfont
  \scalebox{0.8}{    
    \begin{tabular}{lcccc}
    \toprule
    \multirow{2}{*}{Model}&
    \multicolumn{4}{c}{\texttt{Restaurant-Ext}}\cr  
     \cmidrule(lr){2-5} 
    &$\mathtt{Inform}$&$\mathtt{Success}$&$\mathtt{BLEU}$&$\mathtt{Combined}$\cr
    \midrule    
    \modelp{S} &54.00&0.00&6.42&33.42 \\
    \soloiststeach{} &64.00&18.00&9.34&50.34 \\
    % \soloistoateach{} &62.00&20.00&10.89&51.89 \\
    \slsoloistteach{} & \textbf{68.00} & \textbf{24.00} & \textbf{11.76} & \textbf{57.76}\\
    \soloistthteach{} & 68.50 & 26.00 & 11.88 & 59.13\\
    \bottomrule  
    \end{tabular}
    }
  \end{threeparttable}
%   \caption{Evaluation results on task definition extensions. \modelp{S} denotes \soloist{} trained with synthetic dialogs constructed from 5 training dialogs in \texttt{Restaurant} domain. \soloiststeach{} is fine-tuned  with 10 dialogs in \texttt{Restaurant-Ext} provided via machine teaching. \slsoloistteach{}, \soloistthteach{} are refined using 20 real (unlabeled) human-bot dialog logs collected after deployment. (Difference in mean is significant with p<0.01 based on Combined.)}
  
  \caption{Automatic evaluation results on task definition extensions. (Difference in mean is significant with p<0.01 based on Combined.)}
  \label{tab:domain_adapt}
\end{table}

\paragraph{Setup.} We follow the domain extension experiment setting in \citet{lipton2018bbq} to assess the ability of \slsoloist{} to quickly handle task definition extensions. We extend existing \texttt{Restaurant}, denoted as \texttt{Restaurant-Ext}, with additional functions by introducing 4 new slots, \ie \textit{[restaurant\_dish]}, \textit{[value\_price]}, \textit{[start\_time]}, \textit{[end\_time]} in added dialog turns (in Appendix \ref{sec:task_extension}), and corresponding values for each DB entry (in Appendix \ref{sec:ext_db}). The first slot is about the restaurant's signature dish, and the last three are related to delivery service. We leverage Conversational Learner (CL) \cite{shukla2020conversation}, a practical machine teaching tool, to visualize and select dialogs for constructing training examples on the \texttt{Restaurant-Ext} domain by providing corrections and introducing new slots. Finally, 10 examples are obtained through machine teaching for training, 50 for validating and 50 for testing. We fine-tune the dialog model \modelp{S} and the previously trained reward model\footnote{The reward model used for obtaining \slsoloist{} in the Table \ref{tab:results_4task}. It is trained with 5 labeled dialogs in the train set.}, using 10 corrected dialogs, resulting the agent denoted as \soloiststeach{}. Then, \soloiststeach{} is deployed to converse with real human to collect 20 real (unlabeled) human-bot dialog logs, which are then used to refine itself, resulting in \slsoloistteach{}. To better show the effectiveness of the reward model in \slagent{}, we also report the result of \soloistthteach{}, which is refined using the turn-level human feedback score.

\paragraph{Results.} The evaluation results are presented in Table \ref{tab:domain_adapt}. We observe that \modelp{S} has zero success rate, which is predictable as it does not have any knowledge of the new functions. \soloiststeach{} outperforms the baseline by 17 points in terms of $\mathtt{Combined}$ score, which exhibits the effectiveness of machine teaching for handling new functions. \slsoloistteach{} lifts the $\mathtt{Combined}$ score by approximately 7 points, achieving comparable performance with \soloistthteach{}. The results show that \slsoloistteach{} can adapt to new tasks and continually improve itself by automatically learning from the interactions, revealing, with minimum annotations from machine teaching, \slagent{} enables flexible adaptations to new functions.
% (\ie refining using turn-level human feedback score)
\subsection{Interactive Human Evaluation}

\begin{table}[t]
  \centering
  \begin{threeparttable}
  \fontsize{8}{9}
  \selectfont
  \scalebox{0.95}{    
    \begin{tabular}{lccccc}
    \toprule
    \multirow{2}{*}{Model}&
    \multicolumn{5}{c}{\texttt{Restaurant}}\cr  
     \cmidrule(lr){2-6} 
    &$\mathtt{SR\ w/o\ g}$&$\mathtt{SR\ w/\ g}$&$\mathtt{Under.}$&$\mathtt{Appr.}$&$\mathtt{Turns}$\cr
    \midrule
    \modelp{S} &31.82 & 29.54 & 3.86 & 4.13& 10.00\\
    \soloistoa{} &33.42 & 30.86 & 3.89 & 4.12& 9.97\\
    \slsoloist & \textbf{43.10} & \textbf{36.21} & \textbf{3.97} &\textbf{4.13} &\textbf{9.89}\\
    % \soloistth{} & 49.35 & 39.44 & 3.98 & 4.15 & 9.86 \\
    \bottomrule  
    \end{tabular}
    }
  \end{threeparttable}
  \caption{Human evaluation results. SR w/o g: Success rate without grounding, SR w/ g: Success rate with grounding, Under.: Understanding score, Appr.: Appropriateness score.}
  \label{tab:human_evaluation}
  \vspace{-2mm}
\end{table}

% \paragraph{Setup.} Corpus-based evaluation is conducted using automatic evaluation metrics, which are rough proxies for agent response quality. Furthermore, automatic evaluation results may not adequately reflect the capability of dialog systems for helping users complete tasks in the real world, as real user inputs are more dynamic, complex, even with noise. Therefore, we conduct human evaluations to evaluate the performance of \modelp{S}, \soloistoa{}, \slsoloist{} interacting with human users, following the evaluation protocol in DSTC9 track 1 challenge \cite{gunasekara2020overview}, with 100 dialogs gathered for analysis, respectively. 

\paragraph{Setup.} We conduct human evaluations to evaluate the performance of \modelp{S}, \soloistoa{}, \slsoloist{} interacting with human users, following the evaluation protocol in DSTC9 track 1 challenge \cite{gunasekara2020overview}, with 100 Turkers involved and 100 dialogs gathered for analysis, respectively. 

\paragraph{Results.} The human evaluation results on \texttt{Restaurant} domain are presented in Table \ref{tab:human_evaluation}. The results show that \slsoloist{} outperforms \modelp{S}, \soloistoa{} over all the metrics, which are consistent with the automatic evaluation results. The significant improvement on two success rate metrics, especially success rate with grounding, verifies the effectiveness of the reward model in \slagent{} for refining the dialog agent after deployment without additional human annotations, as it more adequately reflects the system's capability of completing tasks in real scenarios. Two interactive examples are in Appendix \ref{sec:int_example}.

\subsection{Ablation Study}

\begin{table}[!t]
  \centering
  \begin{threeparttable}
  \fontsize{8}{9}
  \selectfont
    \begin{tabular}{lcccc}
    \toprule
    \multirow{2}{*}{Reward model}&
    \multicolumn{4}{c}{\texttt{Restaurant}}\cr  
     \cmidrule(lr){2-5} 
    &$\mathtt{Inform}$&$\mathtt{Success}$&$\mathtt{BLEU}$&$\mathtt{Combined} $\cr
    \midrule
    \gpt &67.00&41.50&9.30&63.55 \\
    \bert &68.00&42.50&9.55&64.80 \\
    \bertlarge &66.00&44.00&11.09&66.09\\
    \roberta &72.00&45.00&9.23&67.73\\
    \robertalarge &69.50&46.50&10.20&68.20\\
    \slsoloist & \textbf{75.00} & \textbf{44.50} & \textbf{10.60} &\textbf{70.35}\\
    %\slsoloist$^{*}$ & 71.50 & 47.00 & 11.79 & 71.04 \\
    \bottomrule  
    \end{tabular}
  \end{threeparttable}
%   \caption{Ablation study results on using different PLMs for reward models in \texttt{Restaurant} domain. The first five rows indicate evaluation results of fine-tuned \gpt{}, \bert{}, \bertlarge{}, \roberta{}, \robertalarge{}, respectively. The last row refers to previously reported \slsoloist{}. (Difference in mean is significant with p<0.01 based on Combined.)
  
  \caption{Ablation study results on using different PLMs for reward models. (Difference in mean is significant with p<0.01 based on Combined.)}
  \label{tab:ablation}
  \vspace{-1mm}
\end{table}

% As mentioned before, we implement the reward model of \slagent{} with \gpt{} using multi-task training objective (\ie quality prediction task and enhanced generation task). 
\paragraph{Impact of different PLMs for reward models.} We conduct ablation studies on \texttt{Restaurant} domain to analyze the influence of choosing different PLMs and multi-task training objective on the reward model. We choose several popular PLMs including \bert{} \cite{devlin2018bert} and \roberta{} \cite{liu2019roberta}. Note that all the models share the same pre-training and fine-tuning procedure, except that \bert{} and \roberta{} are trained with quality prediction task while \slsoloist{} is optimized using multi-task learning. We show in Table \ref{tab:ablation} that \roberta{} performs better than \bert{}. \gpt{} (on which \slsoloist{} is built) trained with single quality prediction task, yields significantly worse performance than other methods. We speculate that bidirectional Transformer encoder enables \bert{} and \roberta{} to capture richer context information. \slsoloist{} achieves consistent performance improvements over all the metrics, showing the effectiveness of multi-task learning for the reward model.

\section{Conclusion}
In this paper, we propose a new research problem \ie how to enable task bots to automatically adapt themselves to changing environments by learning from interactions with minimum or zero human annotations. In addition, we propose \slagent{}, a novel self-learning framework. We verify its effectiveness on automatically adapting to changing environments on four dialog tasks by learning from the unlabeled human-bot dialog logs via reinforcement learning with an incorporated pre-trained reward model. As for future work, there are more ways that a task bot could learn to improve itself, \eg during machine teaching, human experts could provide not only correct labels but also feedback in natural language. We leave the theme of effective machine teaching to future work.
% As for future work, there are more ways that a task bot could learn to improve itself, \eg collecting user inputs with diverse language patterns to improve the task bots to identify diverse user inputs. We leave the theme to future work.
\section{Ethical Considerations}
During the collection, annotation and evaluation procedure of the human-bot dialog logs, all involved Amazon Mechanic Turkers and human annotators have been informed of the research purpose in advance, and any of their privacy will not be disclosed or violated during the research period. All other used datasets are open-sourced datasets. In summary, we abide by all research ethics.
\section{Acknowledgements}
This research is affiliated with the CUHK MoE-Microsoft Key Laboratory for Human-centric Interface Technologies. The project is partially sponsored by a grant from the HKSAR Research Grants Council General Research Fund (project number 14207619). In addition, we would like to thank Xixin Wu, Yifei Yuan, Kun Zhang, Kun Li and Jingyan Zhou in particular for their insightful comments.
\bibliography{slsoloist}
\bibliographystyle{acl_natbib}
\newpage

\appendix

\section{RL Refining Algorithm}
\label{sec:alg}
\begin{algorithm}[htb] 
\caption{Self-learning-based RL refining framework.} 
\label{alg:Framwork} 
\begin{algorithmic}[1] %这个1 表示每一行都显示数字
\REQUIRE ~~\\ %算法的输入参数：Input
Training examples $\mathcal{D}$ in the form of dialog turns;\\
Trained agent with dialog model $p_{\boldsymbol{\theta_{D}}}(\boldsymbol{x})$ and reward model $p_{\boldsymbol{\theta_{R}}}(\boldsymbol{x})$.
\ENSURE ~~\\ %算法的输出：Output
Refined agent with updated dialog model $p_{\boldsymbol{{\theta}^{*}_{D}}}$.
\WHILE {not converged}
% \STATE Randomly sample a dialog turn, i.e. token sequences of dialog history $\boldsymbol{s}$ and response $\boldsymbol{r}$;
\STATE Randomly sample a dialog turn, i.e. token sequences of dialog history $\boldsymbol{s}$;
\STATE Run dialog model $p_{\boldsymbol{\theta_{D}}}$ on dialog history $\boldsymbol{x}=(\boldsymbol{s})$ to generate belief state $\boldsymbol{\hat{b}}$;
\STATE Retrieve DB state $\boldsymbol{\hat{c}}$ from a database using generated belief state $\boldsymbol{\hat{b}}$;
\STATE Sample corresponding response $\boldsymbol{r}$ based on dialog history $\boldsymbol{s}$, belief state $\boldsymbol{\hat{b}}$ and DB state $\boldsymbol{\hat{c}}$;
\STATE Use the reward model to predict the quality of the belief state and response with reward score, \\ $R\boldsymbol{(s, \hat{b},\hat{c},r)}$;
\STATE Calculate the loss according to Equation \ref{eqa:RL_ob};\\
\STATE Update the parameters of the dialog model, \\ $\boldsymbol{\theta_{D}} \leftarrow \boldsymbol{\theta_{D}} + \alpha \nabla_{\boldsymbol{\theta_{D}}} \mathcal{L}_{\boldsymbol{\theta_{D}}}$.
\ENDWHILE
\end{algorithmic}
\end{algorithm}

\section{Implementation Details}
\label{sec:imple_details}
\label{sec:ext_db}
\begin{figure}[h]
\centering
\includegraphics[width=0.95\columnwidth]{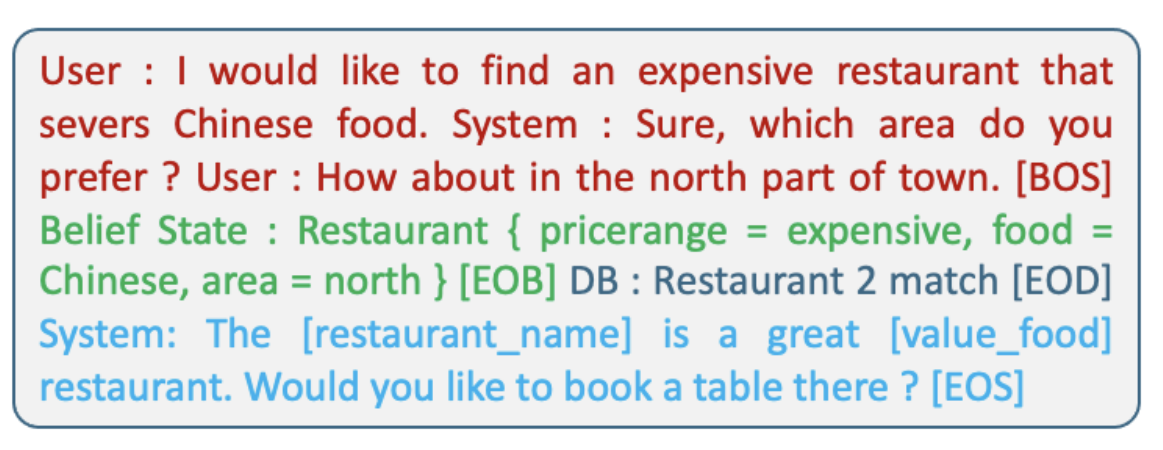}
\caption{Illustration of the training example, \ie the processed dialog turn in the training data.}
\label{fig:dialog_turn}
\end{figure}
To construct training examples as shown in Figure \ref{fig:dialog_turn}, we tokenize the dialog turn sequence using byte pair encodings \cite{sennrich2015neural} and delexicalize responses by replacing slot values with corresponding special slot tokens \cite{lei2018sequicity}. We conduct experiments with several Transformer-based models and \gpt{} \cite{radford2019language} (enhanced with auxiliary generation tasks) shows better performance than others. Therefore, we implement proposed reward model based on Huggingface Pytorch Transformer \cite{wolf2020transformers} using \gpt{}-117M. We pre-train reward model for 10 epochs using Schema dataset \cite{rastogi2019towards}, which contains 22,825 dialogs in 17 domains. The reward model is pre-trained on two 24G Nvidia P40 with a mini-batch of 8 and learning rate of 5e-5, using Adam optimizer \cite{kingma2014adam}, where the training examples are truncated or padded to the max length of 500.

We fine-tune the pre-trained reward model and dialog model (\ie pre-trained \soloist{}) for 20 epochs with limited number of labeled task-specific dialogs for new tasks. During refinement, top-p is selected as 0.5 for all models. We perform gradient clipping with the max norm as 1 for learning model parameters, with the batch size as 1 and learning rate as 5e-6. The dialog model is refined on a single 24G Nvidia P40 until converging on the validation set. During testing, Nucleus filtering is also used for decoding with top-p as 0.5.

\section{Experimental Details}
\label{sec:exper_details}
\begin{figure*}[ht]
\centering
\includegraphics[width=0.95\textwidth]{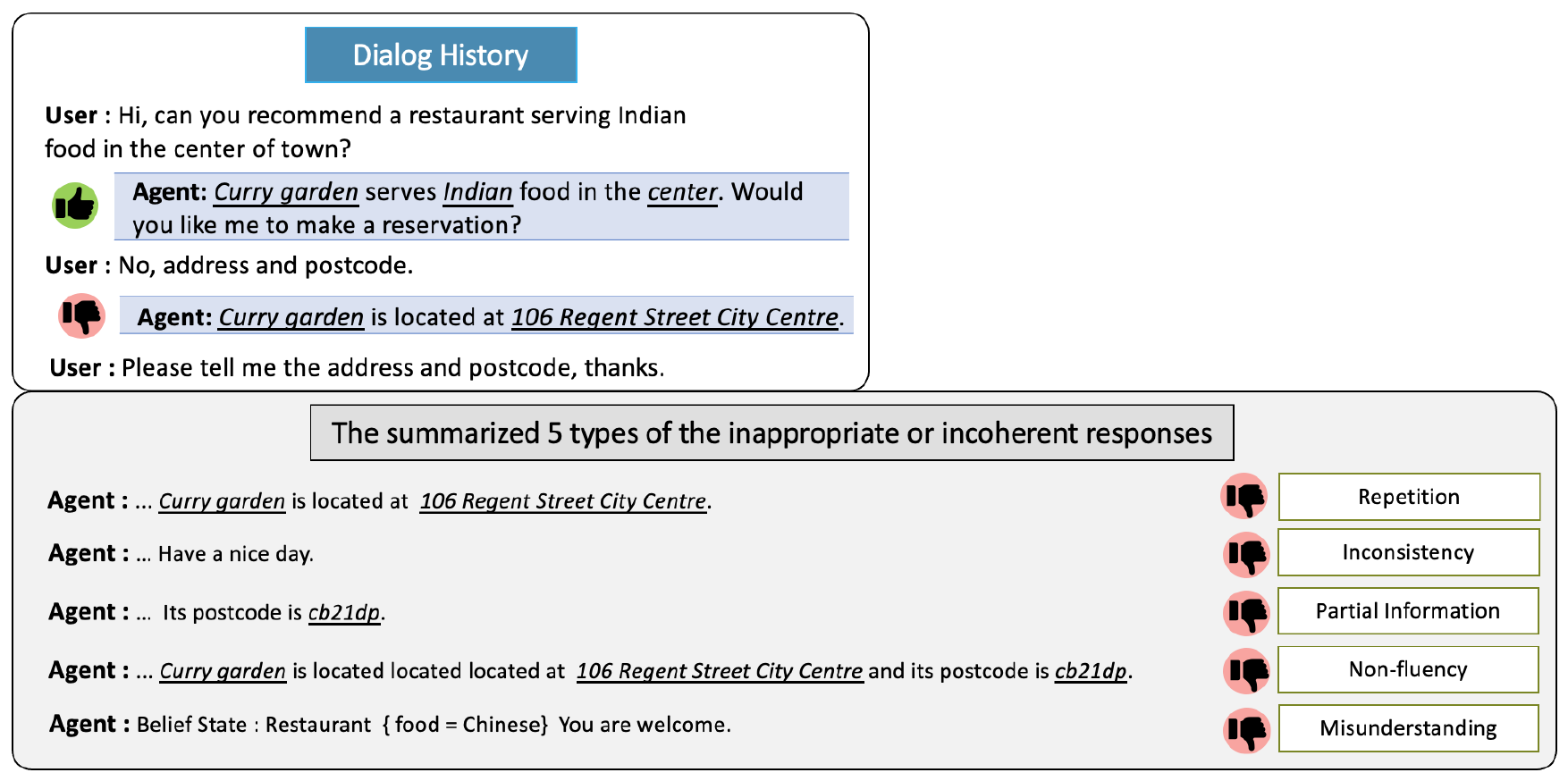} % Reduce the figure size so that it is slightly narrower than the colu
\caption{The summarized 5 types of dialog turns that have inappropriate or incoherent responses. (a) Dialog history (top). (b) 5 types of the inappropriate or incoherent responses (bottom).}
\label{fig:reward}
\vspace{-5mm}
\end{figure*}

To demonstrate the effectiveness of \slagent{}, we use \soloist{} as the dialog model to compare the performance of different methods, since existing state-of-the-art task-oriented dialog models share similar input-output pairs and training objectives as \soloist{}. (We report the results in mean of 5 runs with 5 different seeds.)
$(\RN{1})$ To obtain \modelp{S}, we implement the synthetic dialog construction method by exhausting DB values. For each dialog turn of the 5 labeled dialogs, we randomly sample five DB values from the database to replace the original slot values.
$(\RN{2})$ To obtain \soloistparg{}, we use the Transformer-based machine translation checkpoints (English-German, German-English) \cite{edunov2018understanding} to generate 10 paraphrased user utterances for each dialog turn of the 5 labeled dialogs (based on the empirical analysis of translation quality). Then we use these annotated data (with paraphrased user utterances) to train \modelp{S} for obtaining \soloistparg{}.
$(\RN{3})$ To obtain \soloistoa{}, we use the method described in Section \ref{fig:reward} to construct successful dialogs and failed dialogs. For successful dialogs, we use the original 5 labeled dialogs, and the dialogs containing paraphrased user utterances. To construct the failed dialogs, we randomly select 2-3 dialog turns in each dialog and corrupt responses according to the negative example construction method in Section \ref{reward_model}. Then we use these annotated dialogs to train the session-level reward model of \cite{su2016line}. When testing the performance in the simulated setting, we refine the \modelp{S} with fully correct dialogs and dialogs containing corrupted responses. To achieve better performance, we largely query for session-level human feedback score in both simulated setting and real-scenario setting.

\section{Simulated Human-Bot Corpora Construction}
\label{sec:simu_hb}
The unlabeled simulated human-bot corpora is constructed as follows: $(\RN{1})$ we remove belief state annotations; $(\RN{2})$ we add negative examples by corrupting responses according to the negative example construction method in Section \ref{reward_model}. We will release the simulated human-bot corpora for reproducible research. Note that directly replacing the belief states and responses with the generated ones is trivial. However, such approach cannot imitate realistic human-bot interactions. As the user utterances are strictly fixed, ``users cannot react to the agent responses accordingly and appropriately''. Therefore, we also conduct experiments through conversing with real users in the real-scenario setting and demonstrate the results in Table \ref{tab:policy_improve}. Furthermore, building a user simulator is inapplicable in our changing environment setting. $(\RN{1})$ It is difficult to build reliable user simulators. Building agenda-based user simulators requires sophisticated human expertise for designing rules. $(\RN{2})$ Building model-based user simulators requires sufficient labeled data. Furthermore, model-based user simulators merely imitate expert behaviors in the training corpus, cannot provide user behaviors that are unseen from task bots.

\section{Policy Improvement}
\label{sec:policy_impr}
\begin{table}[!t]
  \centering
  \begin{threeparttable}
  \fontsize{8}{9}
  \selectfont
    \begin{tabular}{lcccc}
    \toprule
    \multirow{2}{*}{Model}&
    \multicolumn{4}{c}{\texttt{Restaurant}}\cr  
     \cmidrule(lr){2-5} 
    &$\mathtt{Inform}$&$\mathtt{Success}$&$\mathtt{BLEU}$&$\mathtt{Combined}$\cr
    \midrule
    \modelp{S} & 62.50&41.50&7.33&59.33\\
    \slsoloist & 75.00&44.50&10.60&70.35\\
    \slsoloist$_{+20}$ & \textbf{75.00}&\textbf{52.00}&\textbf{11.89}&\textbf{75.39}\\
    \bottomrule  
    \end{tabular}
  \end{threeparttable}
  \caption{End-to-end evaluation results of Policy Improvement in the \texttt{Restaurant} domain. \slsoloist$_{+20}$ refer to continually refining with 20 real (unlabeled) human-bot dialogs based on \slsoloist{} (reported in Table \ref{tab:results_4task}).}
  \label{tab:policy_impr}
\end{table}

\paragraph{Policy Improvement Setup.} To demonstrate the effectiveness of \slagent{} for continually learning from collected human-bot dialog logs, we deploy \slsoloist{} online and recruit human users to converse with it to achieve the assigned user goal. We collect 20 real human-bot dialog logs to refine \slsoloist{}, resulting in the agent \slsoloist$_{+20}$. (When refining the \slsoloist{}, we do not use the knowledge about the user's goal. The response quality is judged by the reward model in \slsoloist{}.)

\paragraph{Policy Improvement Results.} The evaluation results on \texttt{Restaurant} are shown in Table \ref{tab:policy_impr}. We observe that \slsoloist$_{+20}$ refined with 20 real (unlabeled) human-bot dialogs outperforms \slsoloist{} by approximately 5 points in terms of $\mathtt{Combined}$ score. We conclude that \slsoloist{} enables continual self-learning after deployment by automatically learning from interactions.

\section{Interactive Example}
\label{sec:int_example}
\begin{figure}[t]
\centering
\includegraphics[width=0.95\columnwidth]{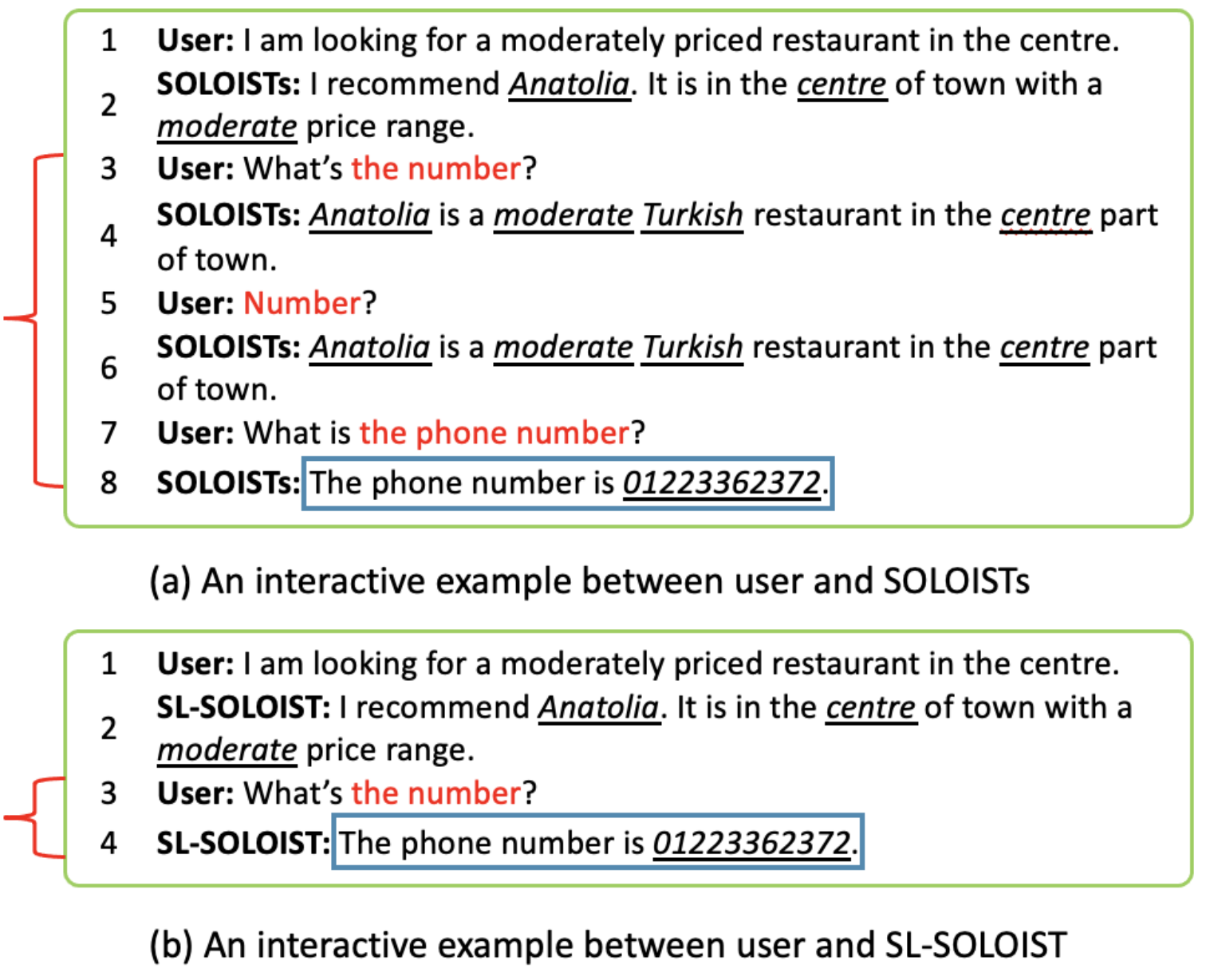}

\caption{Two interactive examples. (a) An interactive example between user and \modelp{S}. (b) An interactive example between user and \slsoloist{}. 
}

\label{fig:case_study}
\end{figure}
%To clearly demonstrate the effectiveness of \slagent{} for handling user behavior variations, we perform a case study to compare the performance of \modelp{S} and \slsoloist{} for completing the same task. 

Figure \ref{fig:case_study} depicts two interactive examples where the same user interacts with \modelp{S} and \slsoloist{} to complete the same task. %The upper example (a) is between user and \modelp{S}, and the lower example (b) is between user and \slsoloist{}. 
We observe that, in the first four dialog turns, the two agents has the same performance and both correctly recommend a satisfied restaurant. However, as shown in Figure \ref{fig:case_study} (a), when user queries about the phone number (``what's the number?'') in the fifth turn, \modelp{S} fails to understand user's intent and generates incoherent response, still trying to provide recommendation. The user has to continually query about phone number in the following consecutive turns. As demonstrated in Figure \ref{fig:case_study} (b), \slsoloist{} correctly provides the phone number, when user first queries about it. Comparing the two examples, we show that \slagent{} enables adapting to unseen user behaviors in an automatic way.
% \section{Training Example}
% \label{sec:dialog_turn}
% \begin{figure}[h]
% \centering
% \includegraphics[width=0.95\columnwidth]{Figures/dialog_turn.pdf}
% \caption{A training example \cite{peng2020soloist}.}
% \label{fig:dialog_turn}
% \end{figure}
\section{An Example of Task Definition Extensions}
\label{sec:task_extension}
\begin{figure}[ht]
\centering
\includegraphics[width=0.95\columnwidth]{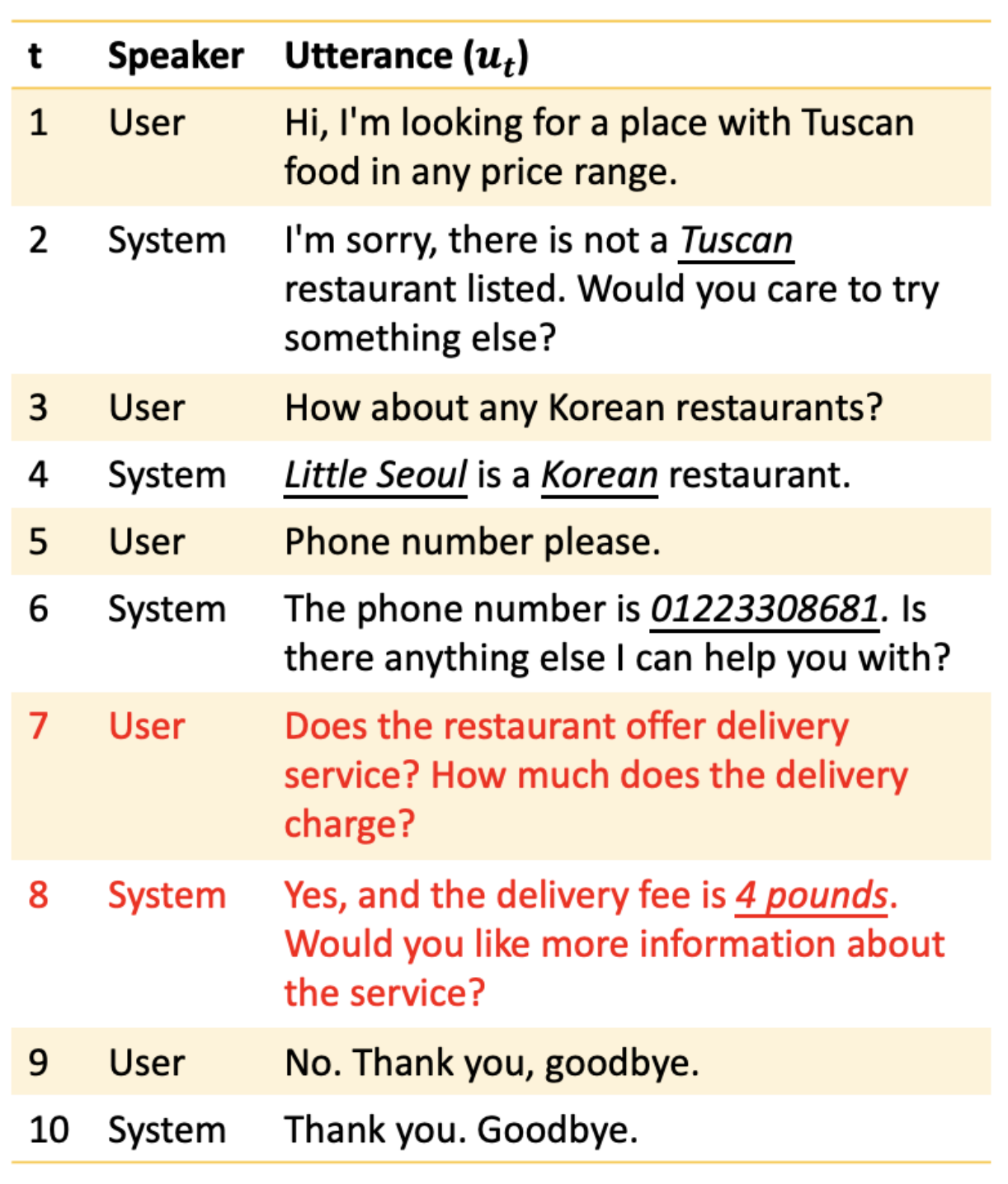}
\caption{An example of task definition extensions. Task bots need learn to provide information about the extended delivery service in additional dialog turns (in Red) as user requirements evolve.}
\label{fig:task_extension}
\end{figure}

Figure \ref{fig:task_extension} depicts an example of task definition extensions.
\section{An Example of \texttt{Restaurant-Ext} DB Entry}
\begin{figure}[ht]
\centering
\includegraphics[width=0.95\columnwidth]{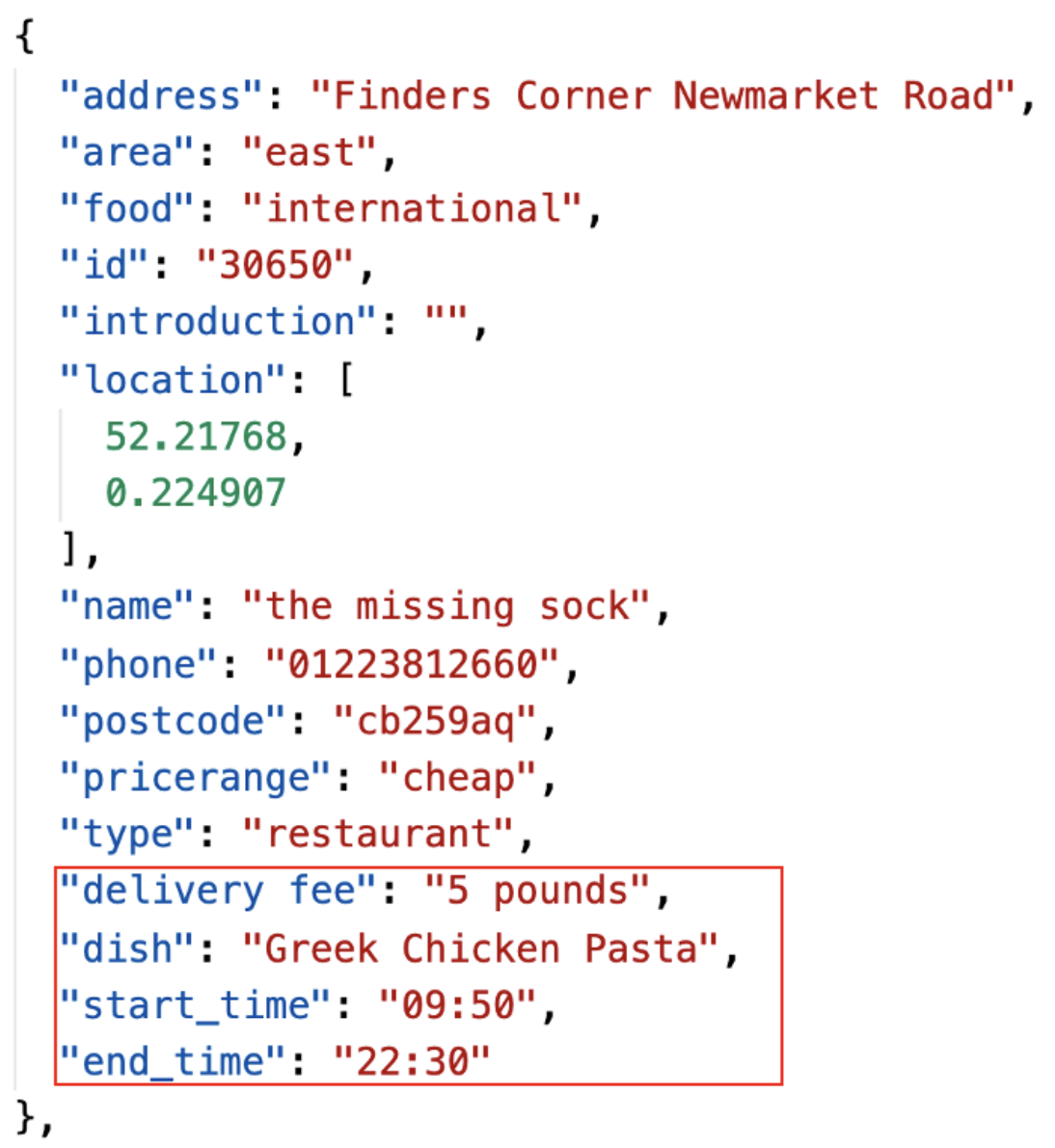}
\caption{An example of \texttt{Restaurant-Ext} DB entry. Newly added DB information about the extended function is in the red square.}
\label{fig:ex_db}
\end{figure}

An example of \texttt{Restaurant-Ext} DB entry is shown in Figure \ref{fig:ex_db}.
\label{sec:ext_db}
\section{Item Examples of the Input Dialog Turn Sequence}
\label{sec:input_example}
\begin{figure*}[ht]
\centering
\includegraphics[width=0.95\textwidth]{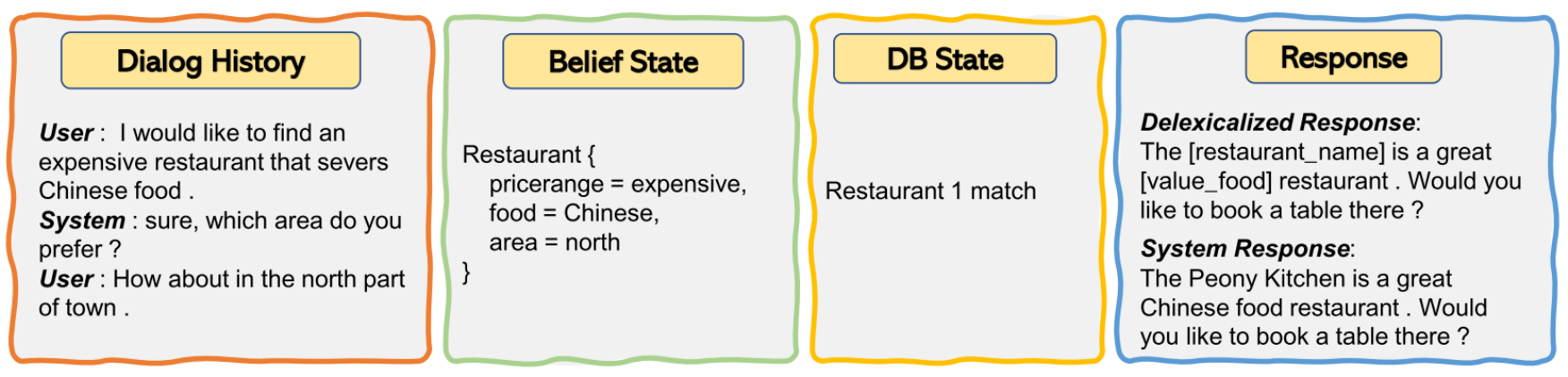}
\caption{Item examples of the input dialog turn sequence for \soloist{}, cited from \cite{peng2020soloist}.}
\label{fig:mt}
\end{figure*}
\section{Negative Example Construction}
\label{sec:neg}

\end{document}